\documentclass{article}
\usepackage{spconf,amsmath,graphicx,hyperref}
\usepackage{amssymb}
\usepackage[ruled,linesnumbered]{algorithm2e}

\title{Neural Nonlinear Shrinkage of Covariance Matrices for Minimum Variance Portfolio Optimization}
\name{Liusha Yang\textsuperscript{1}, Siqi Zhao\textsuperscript{1}, Shuqi Chai
\textsuperscript{2*}\thanks{*Corresponding author}
}
\address{\textsuperscript{1} College of Big Data and Internet, Shenzhen Technology Universiy, China \\
\textsuperscript{2}Shenzhen Research Institute of Big Data, China \\
\texttt{yangliusha@sztu.edu.cn}, ~~\texttt{schai@sribd.cn}}
\begin{document}
%\ninept
%
\maketitle
\begin{abstract}
This paper introduces a neural network–based nonlinear shrinkage estimator of covariance matrices for the purpose of minimum variance portfolio optimization. It is a hybrid approach that integrates statistical estimation with machine learning. Starting from the Ledoit–Wolf (LW) shrinkage estimator, we decompose the LW covariance matrix into its eigenvalues and eigenvectors, and apply a lightweight transformer-based neural network to learn a nonlinear eigenvalue shrinkage function. Trained with portfolio risk as the loss function, the resulting precision matrix (the inverse covariance matrix) estimator directly targets portfolio risk minimization. By conditioning on the sample-to-dimension ratio, the approach remains scalable across different sample sizes and asset universes.
Empirical results on stock daily returns from Standard \& Poor's 500 Index (S\&P500) demonstrate that the proposed method consistently achieves lower out-of-sample realized risk than benchmark approaches. This highlights the promise of integrating structural statistical models with data-driven learning.
\end{abstract}

\section{Introduction}\label{sec:intro}
The classical mean–variance framework of Markowitz~\cite{Markowitz1952} laid the foundation for modern portfolio theory. A key limitation of this approach is its sensitivity to estimation errors in expected returns and the covariance matrix, with the former generally considered harder to estimate reliably~\cite{merton1980estimating,Jagannathan2003}. This has motivated extensive research on improving the global minimum variance portfolio (GMVP), which depends only on covariance estimates.

The sample covariance matrix (SCM) is the most common estimator but performs poorly when the sample size is comparable to the number of assets—a frequent scenario in finance~\cite{el2010high,karoui2013realized,bai2009enhancement,bouchaud2009financial}. To address this, several methods have been proposed: factor models~\cite{chan1999portfolio,fan2008high}, linear shrinkage~\cite{ledoit2003improved,Ledoit&Wolf2004}, nonlinear shrinkage~\cite{ledoit2014nonlinear}, and eigenvalue cleaning techniques~\cite{laloux1999noise,plerou2002random}. Extensions also exist for heavy-tailed and non-stationary returns through robust covariance estimation~\cite{huber1964robust,tyler1987distribution}, and random matrix theory–based robust shrinkage~\cite{couillet2013robust,Hero2011}.

Nonetheless, most estimators optimize generic statistical loss functions, which do not necessarily translate to better portfolio risk outcomes. Moreover, analytically deriving task-specific optimal estimators remains intractable. Machine learning offers a flexible alternative: neural networks can adapt to complex data and be trained with task-oriented objectives. However, end-to-end neural methods often require large datasets and risk ignoring valuable structural prior information. More importantly, the estimation difficulty scales quadratically with the number of assets. 

To overcome these challenges, we propose a hybrid strategy that combines statistical estimation with neural learning. Specifically, we decompose a baseline covariance estimator into eigenvalues and eigenvectors, apply a lightweight neural network (NN) to shrink the eigenvalues, and reconstruct a risk-optimized precision matrix $\hat{\mathbf{C}}_{\mathrm{NN}}^{-1}$. The resulting GMVP achieves consistently lower realized risk than existing approaches such as LW or robust shrinkage estimators~\cite{Ledoit&Wolf2004,Hero2011}.
Our contributions are validated through  empirical analysis on S\&P500 stock daily returns, where the proposed method consistently improves out-of-sample portfolio performance.

\section{Problem formulation}
We consider a time series comprising ${\bf x}_1,...,{\bf x}_n\in\mathbb{R}^N$ logarithmic returns of $N$ financial assets, modeled as
\vspace{-0.1cm}
\begin{align}\nonumber%\label{eq:timeseries}
{\bf x}_t={\pmb\mu}+ {\bf C}_N^{1/2}{\bf y}_t, ~~~t=1,2,...,n,
\vspace{-0.1cm}
\end{align}
where ${\pmb\mu}\in\mathbb{R}^N$ is the mean vector of the asset returns, ${\bf C}_N\in\mathbb{R}^{N\times N}$ is positive definite and ${\bf y}_t\in\mathbb{R}^N$ is a zero mean unitarily invariant random vector with $\|{\bf y}_t\|^2=N$. Both ${\pmb\mu}$ and ${\bf C}_N$ are assumed to be time-invariant over the observation period.  

 Let ${\bf h}\in\mathbb{R}^N$ denote the portfolio selection, i.e., the vector of asset holdings in units of currency normalized by the total outstanding wealth, satisfying ${\bf h}^T{\bf 1}_N=1$. In this paper, short-selling is allowed, and thus the portfolio weights may be negative. Then the portfolio variance (or risk) over the investment period of interest is defined as $\sigma^2({\bf h})=E[|{\bf h}^T{\bf x}_t|^2]={\bf h}^T{\bf C}_N{\bf h}$ \cite{Markowitz1952}. Accordingly, the GMVP selection problem can be formulated as the following quadratic optimization problem with a linear constraint:
 \vspace{-0.3cm}
\[\min\limits_{\bf h}~~~\sigma^2({\bf h})~~~~~
{\rm s.t.}~~{\bf h}^T{\bf 1}_N=1.
\vspace{-0.2cm}
\]
This has the well-known solution $${\bf h}_{\rm GMVP}=\frac{{\bf C}_N^{-1}{\bf 1}_N}{{\bf 1}_N^T{\bf C}_N^{-1}{\bf 1}_N}$$
%\begin{align} \nonumber
%{\bf h}_{\rm GMVP}=\frac{{\bf C}_N^{-1}{\bf 1}_N}{{\bf 1}_N^T{\bf C}_N^{-1}{\bf 1}_N},
%\end{align}
and the corresponding portfolio risk is
\vspace{-0.2cm}
\begin{align} \label{eq:theoretical risk}
\sigma^2\left({\bf h}_{\rm GMVP}\right)=\frac{1}{{\bf 1}_N^T{\bf C}_N^{-1}{\bf 1}_N}.
\vspace{-0.2cm}
\end{align}
Here, (\ref{eq:theoretical risk}) represents the theoretical minimum portfolio risk bound, attained upon knowing the covariance matrix ${\bf C}_N$ exactly. In practice, ${\bf C}_N$ is unknown, and instead we form an estimate, denoted by $\hat{\bf C}_N$. Thus, the GMVP selection based on the plug-in estimator $\hat{\bf C}_N$ is given by
%$\hat{\bf h}_{\rm GMVP}=\frac{\hat{\bf C}_N^{-1}{\bf 1}_N}{{\bf 1}_N^T\hat{\bf C}_N^{-1}{\bf 1}_N}$.
\begin{align} \label{weight}
\hat{\bf h}_{\rm GMVP}=\frac{\hat{\bf C}_N^{-1}{\bf 1}_N}{{\bf 1}_N^T\hat{\bf C}_N^{-1}{\bf 1}_N}.
\end{align}

The quality of $\hat{\bf h}_{\rm GMVP}$, implemented based on the in-sample covariance prediction $\hat{\bf C}_N$, can be measured by its achieved out-of-sample (or ``realized'') portfolio risk:
\vspace{-0.1cm}
\[
\sigma^2\left(\hat{\bf h}_{\rm GMVP}\right)=\frac{{\bf 1}_N^T\hat{\bf C}_N^{-1}{\bf C}_N\hat{\bf C}_N^{-1}{\bf 1}_N}{({\bf 1}_N^T\hat{\bf C}_N^{-1}{\bf 1}_N)^2}.
\vspace{-0.1cm}\]
The goal is to construct a good estimator $\hat{\bf C}_N$, and consequently $\hat{\bf h}_{\rm GMVP}$, which minimizes this quantity.

Note that, for the naive uniform diversification rule, ${\bf h}=\frac{1}{N}{\bf 1}_N$. This is equivalent to setting $\hat{\bf C}_N={\bf I}_N$, and yields the realized portfolio risk: $\frac{{\bf 1}_N^T{\bf C}_N{\bf 1}_N}{N^2}$. Interestingly, this extremely simple strategy has been shown in \cite{DeMiguel2009} to outperform numerous optimized models and will serve as a benchmark in our work.

\section{Novel Precision Matrix Estimator and Portfolio Design for Risk Minimization}

\begin{figure*}[htbp]
\centering
\includegraphics[width=0.8\textwidth]{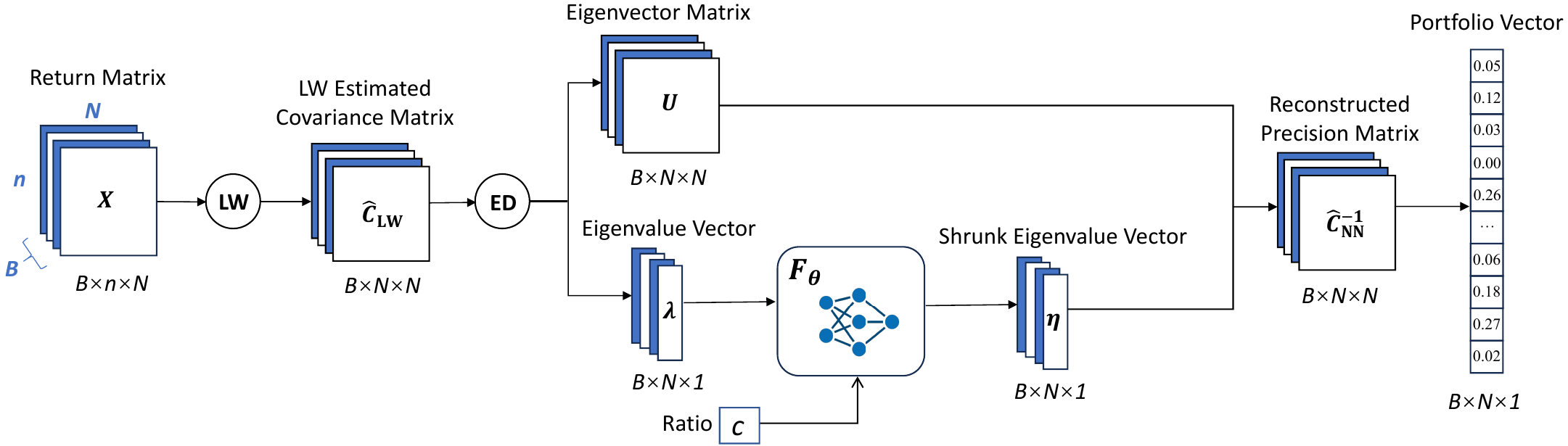}
\vspace{-0.3cm}
\caption{Overall workflow of portfolio optimization with proposed neural network-based nonlinear precision matrix estimator. The ``LW'' represent Ledoit-Wolf covariance matrix estimation. The ``ED'' denotes eigenvalue decomposition.}
\label{fig:algorithm}
\end{figure*}
\begin{figure}[htbp]
\centering
\includegraphics[width=0.45\textwidth]{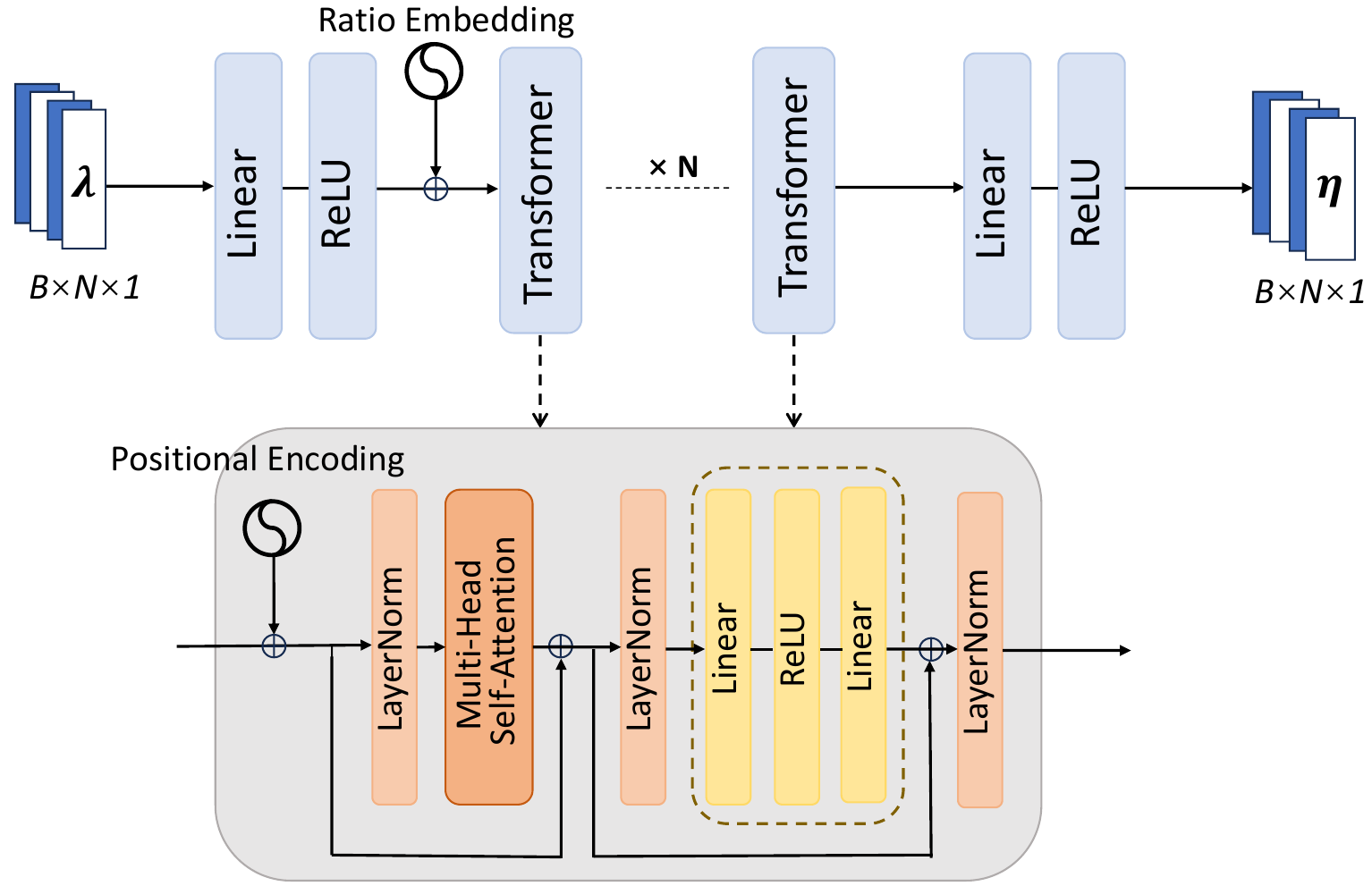}
\vspace{-0.2cm}
\caption{Lightweight transformer-based network architecture.}
\label{fig:network}
\end{figure}

Designing a neural network-based estimator for covariance or precision matrices is inherently challenging. Several obstacles arise: (i) ensuring that the estimated covariance matrix is positive semidefinite—let alone strictly positive definite—is nontrivial; (ii) constructing a network architecture capable of handling matrix-valued inputs of varying dimensions (e.g., different numbers of assets and return samples) is not straightforward; and (iii) since the number of free parameters in the empirical covariance matrix $\mathbf{C}_N$ scales quadratically with $N$, the estimation problem becomes increasingly ill-posed as $N$ grows. To overcome these challenges, we propose a simple yet effective estimation framework that integrates classical statistical techniques with data-driven neural architectures.

Instead of directly estimating the covariance matrix with a neural network, we adopt a hybrid approach that refines existing statistical estimators using neural models, with the objective of reducing portfolio risk. In particular, we build upon the Ledoit–Wolf linear shrinkage estimator \cite{Ledoit&Wolf2004}, which not only improves estimation accuracy compared with the SCM when $N \simeq n$, but also remains invertible in the more challenging regime $N > n$. Fig.~\ref{fig:algorithm} illustrates the overall workflow of the portfolio optimization strategy with the proposed precision matrix estimator. 

\subsection{Precision Matrix Estimation}
The sample covariance matrix is defined as
\vspace{-0.2cm}
\[
{\bf S}_N=\frac{1}{n}\sum_{t=1}^n\tilde{\bf x}_t\tilde{\bf x}_t^T,
\vspace{-0.2cm}
\]
where $\tilde{\bf x}_t={\bf x}_t-\frac{1}{n}\sum_{i=1}^n{\bf x}_i$.  
The LW linear shrinkage covariance matrix estimator is given by a convex combination of the sample covariance matrix and the identity matrix:
\vspace{-0.2cm}
\begin{align}\label{LW}
\hat{\bf C}_{\rm LW}=\rho\mu{\bf I}_N+(1-\rho){\bf S}_N,
\vspace{-1cm}
\end{align}
with $\mu=\frac{1}{N}{\rm tr}[{\bf S}_N]$, $\rho=\dfrac{a^2}{b^2}$, 
$a^2=\|{\bf S}_N-\mu{\bf I}_N\|_{\rm F}^2$,  
$b^2=\tfrac{1}{n}\sum_{t=1}^n\|{\bf x}_t{\bf x}_t^T-{\bf S}_N\|_{\rm F}^2$,  
where $\|\cdot\|_{\rm F}$ denotes the Frobenius norm.

Introducing the eigen-decomposition
\vspace{-0.2cm}
\[
    \hat{\mathbf{C}}_{\mathrm{LW}}
    = \sum_{i=1}^N \lambda_i \mathbf{u}_i \mathbf{u}_i^{\top},
    \vspace{-0.2cm}
\]
where $\lambda_i$ denotes the $i$-th largest eigenvalue and $\mathbf{u}_i$ its associated eigenvector, 
we seek to design a neural network-based precision matrix estimator of the form
\vspace{-0.2cm}
\begin{align}\label{eq:pre_est}
    \hat{\mathbf{C}}_{\rm NN}^{-1}
    = \sum_{i=1}^N \eta_i \mathbf{u}_i \mathbf{u}_i^{\top},
\end{align}
where $\eta_i\geq 0$ is a nonlinear eigenvalue shrinkage function of  
$\boldsymbol{\lambda} = [\lambda_1,\ldots,\lambda_N]$.  
Here the eigenvectors of $\hat{\bf C}_{\rm LW}$ are kept fixed, while the mismatch between ${\bf u}_i$ and the true population eigenvectors is implicitly accounted for by the learned coefficient $\eta_i$.

It is important to emphasize that our goal is to estimate the precision matrix directly, as the portfolio weights in~\eqref{weight} depend explicitly on it. Rather than first estimating $\mathbf{C}_N$ and subsequently inverting it—which can amplify estimation errors—we construct an estimator that directly targets $\mathbf{C}_N^{-1}$.

\subsection{Neural Eigenvalue Shrinkage}
We employ a neural network to learn the nonlinear mapping 
\vspace{-0.2cm}
\[
    \boldsymbol{\eta} = F_\theta(\boldsymbol{\lambda}, c),
    \vspace{-0.2cm}
\]
where $\boldsymbol{\eta} = [\eta_1,\ldots,\eta_N]$ and $c=\tfrac{N}{n}$.  
The ratio $c$ between the number of assets and the number of samples has been shown to play 
a central role in designing shrinkage functions~\cite{ledoit2004well, donoho2018optimal}.  
Conditioned on $c$, the neural network flexibly adjusts the mapping from $\boldsymbol{\lambda}$ to $\boldsymbol{\eta}$ across various $(N,n)$ configurations.  
This allows the method to handle return matrices ${\mathbf{X}}=[{\bf x}_1,\ldots,{\bf x}_n] \in \mathbb{R}^{N\times n}$ of different sizes.  
Once $\boldsymbol{\eta}$ is obtained, the precision matrix estimator $\hat{\mathbf{C}}_{\rm NN}^{-1}$ is reconstructed by~\eqref{eq:pre_est}, and the corresponding portfolio weight vector is
\vspace{-0.2cm}
\[
    \hat{\mathbf{h}}_{\mathrm{NN}}
    = \frac{\hat{\mathbf{C}}_{\rm NN}^{-1}\mathbf{1}_N}{\mathbf{1}_N^{\top}\hat{\mathbf{C}}_{\rm NN}^{-1}\mathbf{1}_N}.
\]

Learning a direct mapping from data ${\mathbf{X}}$ to the precision matrix ${\mathbf{C}}_{N}^{-1}$ is substantially more difficult than mapping eigenvalues $\boldsymbol{\lambda}$ to shrinkage factors $\boldsymbol{\eta}$, particularly when $N$ and $n$ vary.  
By reformulating the task as an eigenvalue shrinkage problem, we significantly reduce complexity and enable the model to generalize more effectively across different sample sizes and portfolio dimensions.

\subsubsection{Network Architecture}
Learning the nonlinear eigenvalue shrinkage function does not require a complicated neural network. We adopt a lightweight architecture based on multiple Transformer layers, which are well suited to modeling complex dependencies and interactions among eigenvalues, as shown in Fig.~\ref{fig:network}.  
The ratio $c$ is embedded into the input to guide the mapping, ensuring that the learned shrinkage function adapts to diverse $\boldsymbol{\lambda}$ derived from $\hat{\mathbf{C}}_{\mathrm{LW}}$ across different sample regimes. Finally, a ReLU activation in the output layer guarantees $\eta_i \geq 0$.

% For completeness, we also benchmark an end-to-end strategy, referred to as direct weight estimation (DWE), 
% which directly maps ${\mathbf{X}}$ to the portfolio weights ${\mathbf{h}}$ (see Section~\ref{sec:realdata}) with the same network architecture.  
% However, this approach is less effective because it does not leverage structural properties inherent in the optimization problem, 
% making training considerably more challenging. Moreover, the model lacks flexibility in adapting to varying sample sizes, further limiting its practical applicability.

\subsubsection{Training Objective}

The neural network is trained with portfolio risk as the loss function.  
Given the eigenvalue mapping $\boldsymbol{\eta} = F_\theta(\boldsymbol{\lambda}, c)$, 
we construct $\hat{\mathbf{C}}_{\rm NN}^{-1}$ via~\eqref{eq:pre_est} and compute the resulting portfolio weights as
\[
    \hat{\mathbf{h}}_{\mathrm{NN}}
    = \frac{\hat{\mathbf{C}}_{\rm NN}^{-1}\mathbf{1}_N}{\mathbf{1}_N^{\top}\hat{\mathbf{C}}_{\rm NN}^{-1}\mathbf{1}_N}.
\]
The associated portfolio risk is
\begin{align}\nonumber
    \mathcal{L} = \hat{\mathbf{h}}_{\mathrm{NN}}^{\top}\mathbf{C}_{N} \hat{\mathbf{h}}_{\mathrm{NN}}.
\end{align}
Since $\mathbf{C}_N$ is unknown in practice, 
we estimate risk using the out-of-sample variance of portfolio returns: 
\begin{align}\label{eq:loss_out_sample}
    \mathcal{L} = \hat{\mathbf{h}}_{\mathrm{NN}}^{\top}\left(\frac{1}{m}\sum_{t=n+1}^{n+m}
    \tilde{\mathbf{x}}_t \tilde{\mathbf{x}}_t^{\top}\right)\hat{\mathbf{h}}_{\mathrm{NN}},
\end{align}
where $\tilde{\mathbf{x}}_t = \mathbf{x}_t - \frac{1}{m}\sum_{i=n+1}^{n+m}\mathbf{x}_i$ denotes centered returns in the $m$ out-of-sample periods.
The complete training procedure of the proposed method is summarized in Algorithm~\ref{alg}.  

\begin{algorithm}[htbp]
\caption{Training pipeline for shrinkage function $F_\theta$}
\label{alg}
\KwIn{Training samples ${\bf x}_1,\ldots, {\bf x}_n$ with randomly chosen $(N,n)$; validation samples ${\bf x}_{n+1}, \ldots, {\bf x}_{n+m}$; ratio $c$} 
\KwOut{Loss value $\mathcal{L}$}

\BlankLine

Compute $\hat{\bf C}_{\rm LW}$ using~(\ref{LW}) from $\{{\bf x}_1,\ldots,{\bf x}_n\}$\; 

Perform eigen-decomposition: $\hat{\bf C}_{\rm LW} \to \boldsymbol{\lambda}, \{{\bf u}_i\}_{i=1}^N$\;

Apply shrinkage function: $\boldsymbol{\eta} \gets F_\theta(\boldsymbol{\lambda}, c)$\;

Construct inverse covariance: $\hat{\bf C}_{\rm NN}^{-1} = \sum_{i=1}^N \eta_i {\bf u}_i {\bf u}_i^{\!T}$\;

Estimate weights:
$\hat{\bf h}_{\rm NN} = \dfrac{\hat{\bf C}_{\rm NN}^{-1}{\bf 1}_N}
{{\bf 1}_N^{\!T} \hat{\bf C}_{\rm NN}^{-1}{\bf 1}_N}$\;

Compute loss $\mathcal{L}$ with $\{{\bf x}_{n+1},\ldots,{\bf x}_{n+m}\}$ using~(\ref{eq:loss_out_sample}).

\end{algorithm}

\section{Experimental Results}

\subsection{Dataset and training configuration}
We use stocks from the S\&P500 index. Specifically, we collect dividend-adjusted daily closing prices from the Yahoo Finance database and compute continuously compounded (logarithmic) returns for a randomly selected subset of 50 constituents. The dataset spans 3,675 trading days from Jan. 3, 2011 to Jul. 31, 2025 (excluding weekends and public holidays). We split the samples into a training set of 3,275 observations covering Jan. 3, 2011 to Dec. 31, 2023, and a test set of 400 observations covering Jan. 3, 2024 to Jul. 31, 2025.

The number of transformer layers is set to be 6. Training is performed with the Adam optimizer (learning rate $10^{-4}$, batch size $8$).
\vspace{-0.3cm}
\subsection{Baselines}
We compare the performance of $\hat{\mathbf{C}}_{\rm NN}^{-1}$ with several benchmark methods:
1) $\hat{\mathbf{C}}_{\rm SCM}$: the sample covariance matrix;
2) $\hat{\mathbf{C}}_{\rm LW}$: the Ledoit–Wolf shrinkage estimator~\cite{Ledoit&Wolf2004};
3) $\hat{\mathbf{C}}_{\rm C}$: the Chen estimator~\cite{Hero2011};
4) ${\mathbf I}_N$: the identity matrix, corresponding to equally weighted portfolios~\cite{DeMiguel2009};
5) \textbf{Direct Weight Estimation (DWE)}: a neural network with the same architecture as $\hat{\mathbf{C}}_{\rm NN}^{-1}$ but trained to directly output portfolio weights from historical returns, without explicit covariance estimation. It is trained using $N=50$ assets and a window length of $n=60$. At test time, when $n > 60$, the input is truncated to 60 observations, whereas the method is not applicable for $n < 60$.

\begin{figure}[htbp]
\centering
\includegraphics[width=0.35\textwidth]{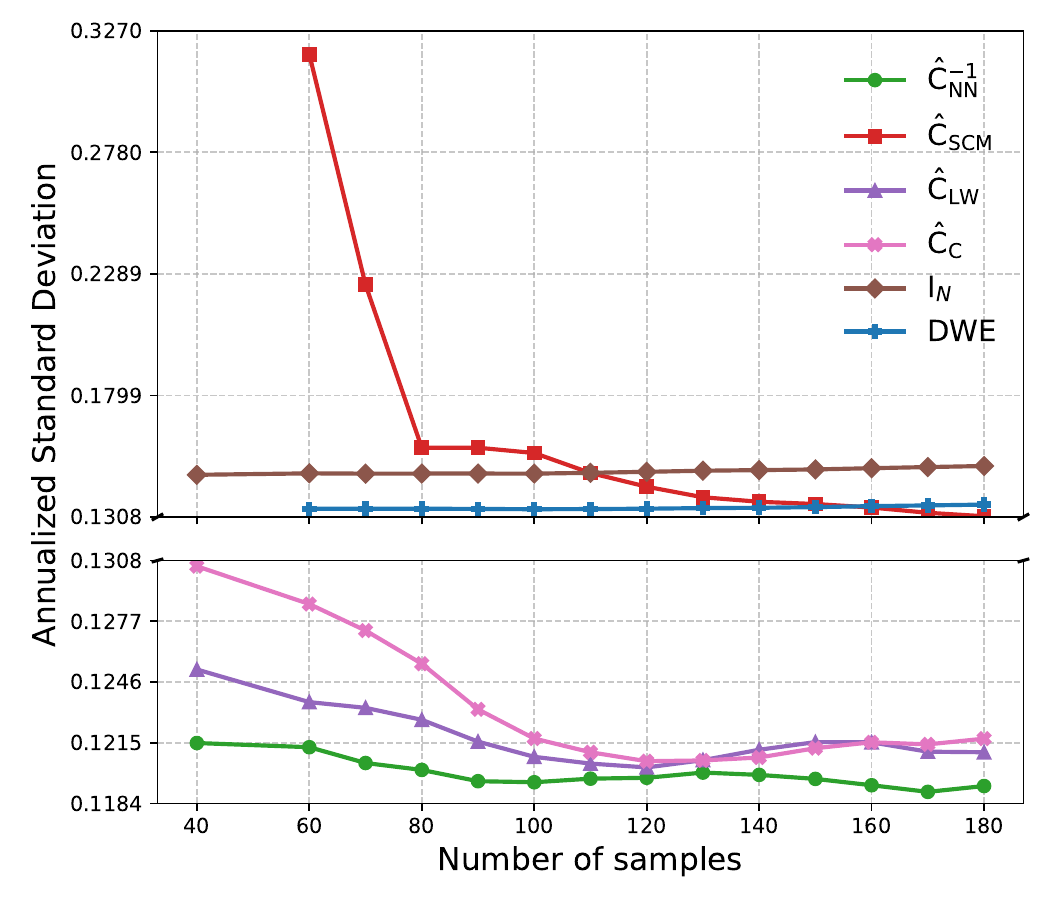}
\vspace{-0.5cm}
\caption{Realized portfolio risks achieved out-of-sample over 400 days of S\&P500
real market data by a GMVP implemented
using different methods.}
\label{fig:risk}
\end{figure}
\vspace{-0.2cm}

% \vspace{-0.2cm}
% \begin{enumerate}
% \itemsep-0.1cm
% \item $\hat{\mathbf{C}}_{\rm SCM}$: the sample covariance matrix;
% \item $\hat{\mathbf{C}}_{\rm LW}$: the Ledoit–Wolf shrinkage estimator~\cite{Ledoit&Wolf2004};
% \item $\hat{\mathbf{C}}_{\rm C}$: the Chen estimator~\cite{Hero2011};
% \item ${\mathbf I}_N$: the identity matrix, corresponding to equally weighted portfolios~\cite{DeMiguel2009};
% \item \textbf{Direct Weight Estimation (DWE)}: a neural network with the same architecture as $\hat{\mathbf{C}}_{\rm NN}^{-1}$ but trained to directly output portfolio weights from historical returns, without explicit covariance estimation. It is trained using $N=50$ assets and a window length of $n=60$. For $n>60$, the input is truncated to 60, and for $n<60$ the method cannot be applied.
% \end{enumerate}

\begin{figure}[htbp]
\centering
\includegraphics[width=0.35\textwidth]{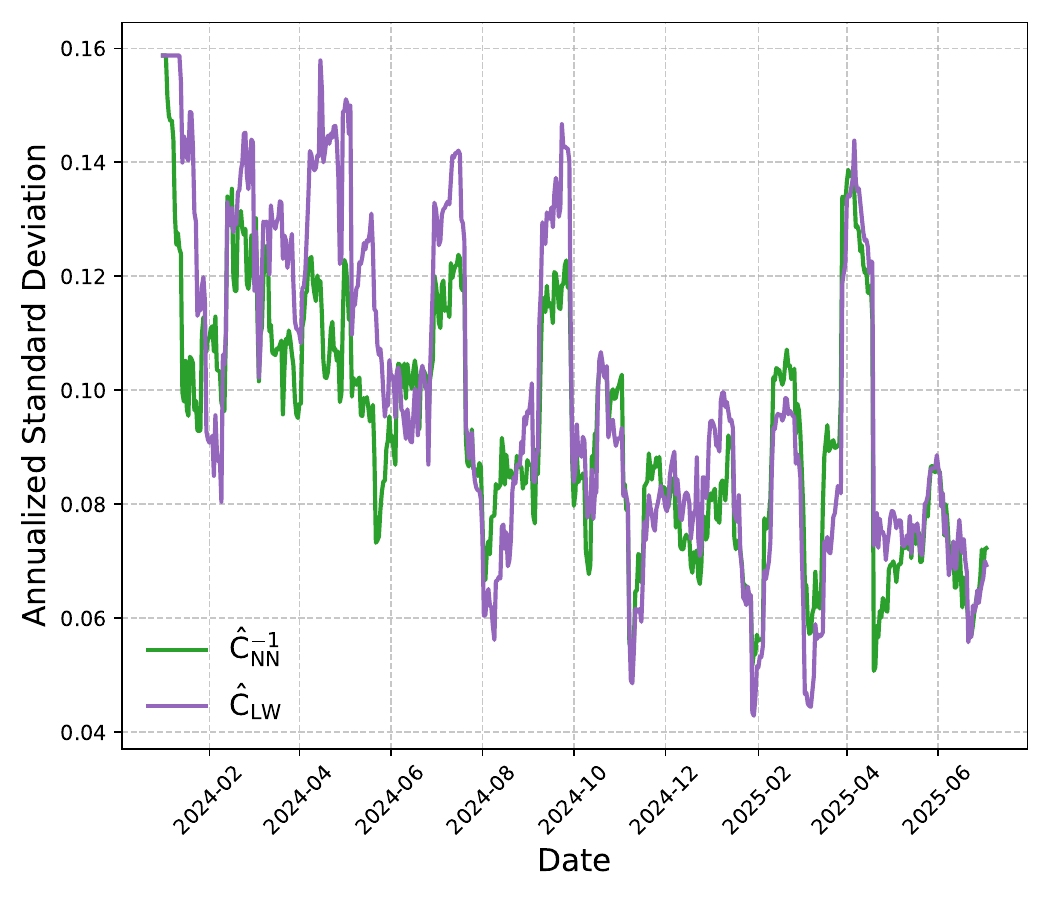}
\vspace{-0.5cm}
\caption{Annualized rolling-window standard deviations of the most recent 40
out-of-sample log returns for the GMVP based on $\hat{\bf C}_{\rm NN}^{-1}$ and $\hat{\bf C}_{\rm LW}$ (N=50, n=100).}
\label{fig:risk2}
\end{figure}

\vspace{-0.2cm}
\subsection{Experimental results}\label{sec:realdata}
We evaluate the out-of-sample performance of different covariance estimators on S\&P500 stock returns using a rolling-window approach. At day $t$, the previous $n$ observations ($t-n$ to $t-1$) form the data window for covariance estimation, from which the GMVP weights $\hat{\bf h}_{\rm GMVP}$ are computed. The resulting portfolio is held for the next 20 days, after which the window shifts forward by 20 days. This procedure continues until the sample ends. Realized risk is measured by the annualized standard deviation of GMVP returns, with multiple training window lengths considered.

Fig.~\ref{fig:risk} shows that the proposed $\hat{\bf C}_{\rm NN}^{-1}$ consistently yields the lowest realized risk across both the $n \leq N$ and $n > N$ regimes, outperforming all benchmarks. The curves for $\hat{\bf C}_{\rm SCM}$ and DWE are absent at $n=40$, since $\hat{\bf C}_{\rm SCM}$ is non-invertible when $n \leq N$ and DWE cannot adapt to input sizes below its training setting ($n=60$).

For finer temporal comparison, we analyze the realized risks of the best two methods $\hat{\bf C}_{\rm NN}^{-1}$ and $\hat{\bf C}_{\rm LW}$ under $n=100$. From 300 out-of-sample portfolio returns, rolling annualized risks are computed using the most recent 40 returns, yielding 261 measurements (Fig.~\ref{fig:risk2}). Results show that $\hat{\bf C}_{\rm NN}^{-1}$ achieves lower risk in 72.1\% of cases, particularly during volatile periods such as Apr.–Jun. 2024.

\section{Conclusion}
This paper has proposed a novel minimum variance portfolio optimization strategy built on a nonlinear shrinkage precision matrix estimator, where the shrinkage function is learned through a neural network and explicitly calibrated to minimize realized portfolio risk. By employing data-driven eigenvalue shrinkage, the method consistently outperforms standard benchmarks in terms of realized risk, as evidenced by empirical results on historical S\&P500 stock returns. A promising direction for future work is to extend the framework to jointly estimate both covariance and expected returns, thereby enabling optimization under alternative objectives such as Sharpe ratio maximization or the classical Markowitz mean–variance formulation.

\clearpage

% References should be produced using the bibtex program from suitable
% BiBTeX files (here: strings, refs, manuals). The IEEEbib.bst bibliography
% style file from IEEE produces unsorted bibliography list.
% -------------------------------------------------------------------------
\bibliographystyle{IEEEbib}
\bibliography{Cited_2}
%\bibliography{strings,refs}
\end{document}